%% file: paper.tex
\documentclass[runningheads]{llncs}
\usepackage{graphicx}
\pdfoutput=1

\usepackage{tikz}
\usepackage{comment} 
\usepackage{amsmath,amssymb} 
\usepackage{color}

\usepackage{pifont}
\newcommand{\cmark}{\ding{51}}%
\newcommand{\xmark}{\ding{55}}%
\newcommand{\etal}{\textit{et al}.}

\begin{document}
\pagestyle{headings}
\mainmatter
\def\ECCVSubNumber{3702}  

\title{Adaptive Video Highlight Detection by Learning from User History} 

\titlerunning{Adaptive Video Highlight Detection by Learning from User History}
%
\author{Mrigank Rochan\inst{1} \and
Mahesh Kumar Krishna Reddy\inst{1} \and
Linwei Ye\inst{1} \and
Yang Wang\inst{1,2}}
\authorrunning{M. Rochan et al.}
%
\institute{University of Manitoba, Winnipeg MB R3T 2N2, Canada \and
Huawei Technologies, Canada\\
\email{\{mrochan,kumarkm,yel3,ywang\}@cs.umanitoba.ca}}
\maketitle

\begin{abstract}
\input abstract.tex
\keywords{Video highlighight detection \and User-adaptive learning}
\end{abstract}

\input intro.tex
\input related.tex
\input approach.tex
\input experiments.tex
\input vsumm.tex
\input conclusion.tex

\clearpage
%
%
\bibliographystyle{splncs04}
\bibliography{rochan_cleaned}
\end{document}

%% file: abstract.tex
Recently, there is an increasing interest in highlight detection research where the goal is to create a short duration video from a longer video by extracting its interesting moments. However, most existing methods ignore the fact that the definition of video highlight is highly subjective. Different users may have different preferences of highlight for the same input video. In this paper, we propose a simple yet effective framework that learns to adapt highlight detection to a user by exploiting the user's history in the form of highlights that the user has previously created. Our framework consists of two sub-networks: a fully temporal convolutional highlight detection network $H$ that predicts highlight for an input video and a history encoder network $M$ for user history. We introduce a newly designed temporal-adaptive instance normalization (T-AIN) layer to $H$ where the two sub-networks interact with each other. T-AIN has affine parameters that are predicted from $M$ based on the user history and is responsible for the user-adaptive signal to $H$. Extensive experiments on a large-scale dataset show that our framework can make more accurate and user-specific highlight predictions.

%% file: intro.tex
\section{Introduction}\label{sec:intro}
There is a proliferation in the amount of video data captured and shared everyday. It has given rise to multifaceted challenges, including editing, indexing and browsing of this massive amount of video data. This has drawn attention of the research community to build automated video highlight detection tools. The goal of highlight detection is to reduce an unedited video to its interesting visual moments and events. A robust highlight detection solution can enhance video browsing experience by providing quick video preview, facilitating video sharing on social media and assisting video recommendation systems. 

\begin{figure}
	\centering
	\includegraphics[width=0.6\textwidth]{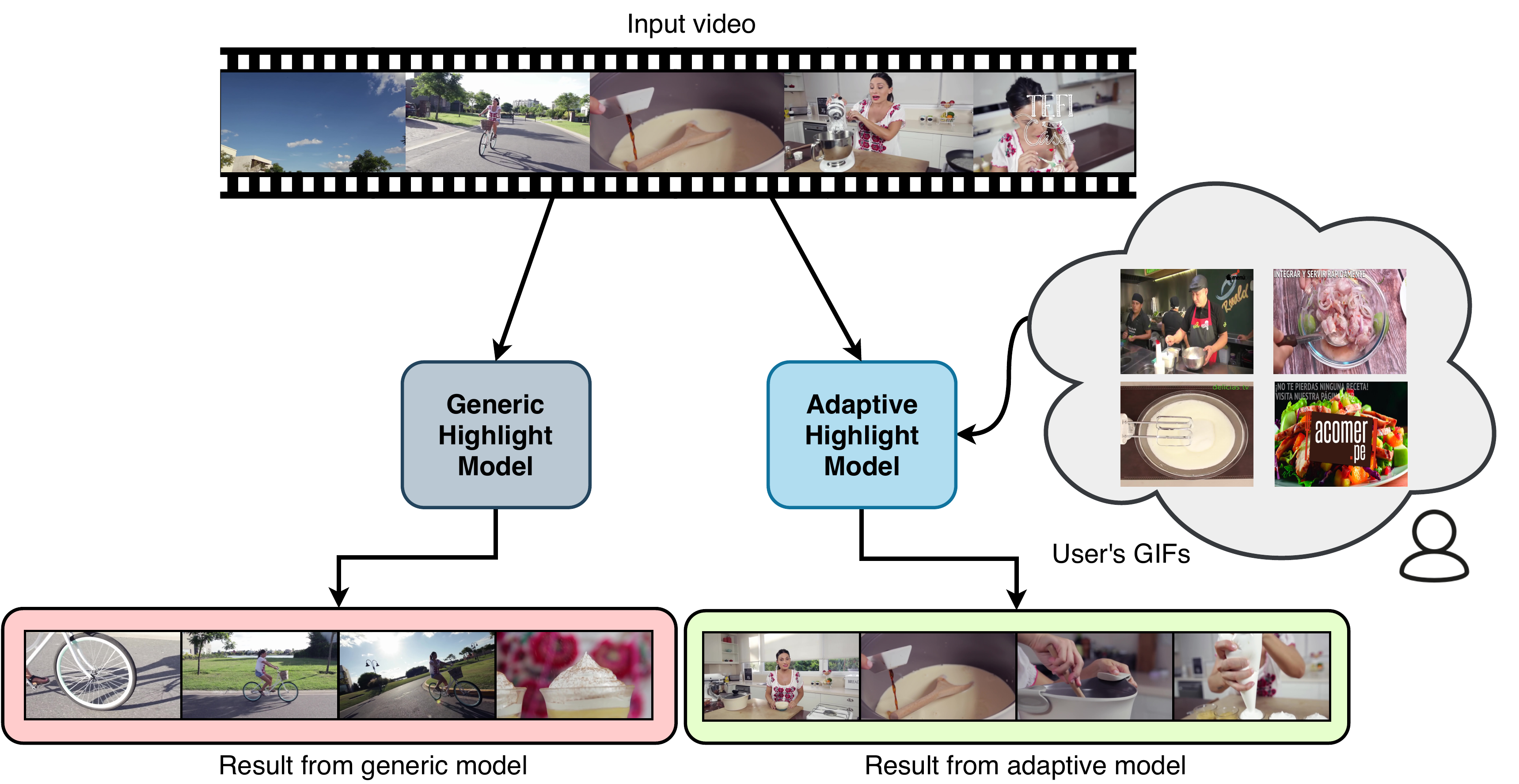}
	\caption{The definition of highlight of a video is inherently subjective and depends on each user's preference. In contrast to a generic highlight detection model, an adaptive highlight detection model (like ours) incorporates a user's previously created highlights (e.g., GIFs from multiple videos) when predicting highlights of an input video. This allows the model to make more accurate and user-specific highlight predictions.}
	\label{fig:intro}
\end{figure}

Even though we have made significant progress in highlight detection, the existing methods are missing the ability to adapt its predictions to users. The main thrust of research in highlight detection has been on building generic models. However, different users have different preferences in term of detected highlights~\cite{gygli18_acmmm,soleymani2015quest}. Generic highlight detection models ignore the fact that the definition of a video highlight is inherently subjective and depends on each individual user's preference. This can greatly limit the adoption of these models in real-world applications. In Fig.~\ref{fig:intro}, we illustrate the subjective nature of highlights. The input video contains events such as cycling, cooking, and eating. A generic highlight detection model mainly predicts the cycling event as the highlight. But if we examine the user's history (previously created highlights by the user), we can infer that this user is interested in cooking scenes. Therefore, a highlight detection model should predict the cooking event as the highlight instead. Motivated by this observation, we propose \textit{an adaptive highlight detection model} that explicitly takes user's history into consideration while generating highlights.

Although a user's visual highlight history can provide a stronger and more reliable cue of their interests than non-visual meta-data \cite{gygli18_acmmm}, there is very limited research on adapting highlight detection using this visual information. To the best of our knowledge, the recent work by Molino and Gygli \cite{gygli18_acmmm} is the only prior work on this topic. Their method considers a user's previously created highlights (available as GIFs\footnote[1]{GIF is an image format with multiple frames played in a loop without sound \cite{gygli16_cvpr}.}) from multiple videos when generating new highlights for that user. They propose a ranking model that predicts a higher score for interesting video segments as opposed to non-interesting ones while conditioning on the user's past highlights (i.e., user's history). However, their method has some limitations. Firstly, it operates at the segment level and samples a fixed number of positive and negative segments from a video for learning. This means that the method does not process an entire video which is essential to capture temporal dependencies shown to be vital in many video understanding tasks \cite{lea17_cvpr,mahasseni17_cvpr,ours18_eccv,zhang16_eccv}. Moreover, it is sensitive to the number of positive and negative samples used in the learning. Secondly, it requires to precompute shot boundaries using a shot detection algorithm \cite{gygli18_shot} for sampling a set of positive/negative video segments. This makes their pipeline computationally complex and expensive. Lastly, their model directly combines user's history features with the features of sampled video segments to predict user-specific highlights. We demonstrate in our experiments that this is not as effective as our proposed model.

In this paper, we introduce a novel user-adaptive highlight detection framework that is simple and powerful. Given an input video for highlight detection and the user's history (highlight GIFs from multiple videos that the user has previously created), our model seamlessly combines the user's history information with the input video to modulate the highlight detection so as to make user-specific and more precise highlight prediction. Our framework consists of two sub-networks: a fully temporal convolutional \textit{highlight detection network} $H$ which produces the highlight for an input video, and a \textit{history encoder network} $M$ that encodes the user's history information. We propose \textit{temporal-adaptive instance normalization} (T-AIN), a conditional temporal instance normalization layer for videos. We introduce T-AIN layers to $H$ where the interaction between the two sub-networks $H$ and $M$ takes place. T-AIN layers have affine transformation parameters that are predicted from $M$ based on the user's history. In other words, $M$ acts a guiding network to $H$. Through the adjustable affine parameters in T-AIN layers, $H$ can adapt highlight predictions to different users based on their preferences as expressed in their histories. Note that our method does not require expensive shot detection. Moreover, it can utilize an entire video for learning instead of a few sampled video segments.

To summarize, the main contributions of our paper are the following. (1) We study user-adaptive highlight detection using user history in the form of previously created highlights by the user. This problem has many commercial applications, but is not well studied in the literature. (2) Different from ranking models \cite{gygli16_cvpr,gygli18_acmmm,xiong19_cvpr,yao2016highlight} commonly used in highlight detection, we are first to employ a fully temporal convolutional model in highlight detection. Our model does not require expensive shot detection algorithm and can process an entire video at once. This makes our proposed model simpler operationally. (3) We propose temporal-adaptive instance normalization layer (T-AIN) for videos. T-AIN layers have adjustable affine parameters which allow the highlight detection network to adapt to different users. (4) We experiment on a large-scale dataset and demonstrate the effectiveness of our approach. (5) We further explore the application of our model as a pre-trained model for video summarization, a task closely related to highlight detection.

%% file: related.tex
\section{Related Work}\label{sec:related}
Highlight detection aims to identify key events in a video that a user is likely to find interesting. Towards this, we exploit the highlights that the user has created in the past. In this section, we discuss the several lines of related work.

Our work is closely related to existing highlight detection methods where the goal is to extract the interesting moments from a long duration video \cite{gygli18_acmmm,truong07_tomm}. Many prior methods~\cite{gygli16_cvpr,jiao2017video,gygli18_acmmm,sun2014ranking,yao2016highlight,yu2018deep} employ a ranking formulation. They learn to score interesting segments of a video higher than non-interesting segments. Our work is particularly related to \cite{gygli16_cvpr,gygli18_acmmm}. Gygli~\etal~\cite{gygli16_cvpr} propose a GIF creation technique using a highlight detection method. The limitation of this method is that it is a generic highlight detector, whereas we propose a model that is capable of making user-specific predictions. Molino and Gygli \cite{gygli18_acmmm} propose a model that takes a user's history as an input to make personalized predictions. Their method is a ranking model that operates on a few sampled positive and negative segments of a video combined with the user's highlight history to learn a personalized model. Different from them, our proposed highlight detection model is convolutional in nature. It can accept an entire video (of any length) and the user's history of different sizes as an input to produce a personalized highlight. Unlike \cite{gygli18_acmmm}, we do not require shot detection as pre-processing. Lastly, we develop a more effective way to leverage user history representation in the network instead of directly concatenating with the input video features~\cite{gygli18_acmmm}.

This paper has also connection with the research in video summarization. Different from highlight detection, video summarization aims to generate a concise overview of a video~\cite{gygli18_acmmm,truong07_tomm}. Early research~\cite{khosla2013large,kim2014reconstructing_cvpr,lee2012discovering,lu2013story,mahasseni17_cvpr,ngo2003automatic_iccv,panda2017collaborative,Song_2015_CVPR,zhou2018_aaai,ours18_eccv,zhang2018_eccv} on summarization mainly develop unsupervised methods. These methods design hand-crafted heuristics to satisfy properties such as diversity and representativeness in the generated summary. Weakly supervised methods also exist that exploit cues from web images or videos~\cite{cai2018weakly,khosla2013large,kim2014reconstructing_cvpr,ours19_cvpr,Song_2015_CVPR} and video category details~\cite{panda2017weakly,potapov14_eccv} for summarization. Supervised methods~\cite{gong14_nips,gygli14_eccv,gygli15_cvpr,ours18_eccv,zhang16_cvpr,zhang16_eccv,zhao2017_acmm} have shown to achieve superior performance due to the fact that they learn directly from annotated training data with human-generated summaries. However, these methods do not consider each user's preference. In theory, it is possible to make user-adaptive predictions by performing user-specific training. But it is expensive in practice as it would result in per-user model \cite{gygli18_acmmm}. However, we learn a single model that can be adapted to different users by incorporating their histories. Moreover, we do not need to retrain the model for a new user. The adaption to a new user only involves some simple feed-forward computation. Lastly, in video summarization, there is some work that focus on personalization but they either use meta-data~\cite{agnihotri2005framework,babaguchi2007learning,jaimes2002learning,takahashi2007user} or consider user textual query~\cite{liu2015multi,sharghi2016query,vasudevan2017query,zhang2018_bmvc} to personalize video summary. Different from them, our method operates on visual features from user video and user's history to make user-specific prediction.

Finally, our approach is partly inspired by recent research in style transfer in images using conditional normalization layers, e.g., adaptive instance normalization~\cite{huang17_cvpr} and conditional batch normalization~\cite{de2017modulating}. Apart from style transfer, both layers have been applied in many other computer vision problems~\cite{brock2018large,chen2018self,karras2019style,liu19_iccv,park19_cvpr,wang2018recovering}. These layers firstly normalize the activations to zero mean and unit variance, then adjust the activations via affine transformation parameters inferred from external data \cite{park19_cvpr}. Since the layers are designed to uniformly modulate the activations spatially, their applications are limited to images. In contrast, in this paper, we propose a normalization layer that uniformly modulates the activations temporally, making it appropriate for video understanding tasks.

%% file: approach.tex
\section{Our Approach}\label{sec:approach}
Given a video $\textbf{v}$, we represent each frame in the video as a $D$-dimensional feature vector. The video can be represented as a tensor of dimension $1 \times T \times D$, where $T$ and $D$ are the number of frames and dimension of frame feature vector of each frame in the video, respectively. $T$ varies for different videos. For an user $U$, let $\mathcal{H}=\{h_1,...,h_n\}$ denotes the \textit{user's history} which is a collection of visual highlights that the user has created in the past.

Given $\textbf{v}$ and $\mathcal{H}$, our goal is to learn a mapping function $G$ that predicts two scores for each frame in $\textbf{v}$ indicating it being non-highlight and highlight. Thus, the final output $S$ is of dimension ${1 \times T\times2}$ for the input video $\textbf{v}$:
\begin{equation}
S=G(\textbf{v},\mathcal{H})= G(\textbf{v},\{h_1,...,h_n\}).
\label{eq:fn_G}
\end{equation}
We refer to $G$ as the adaptive highlight detector.

\subsection{Background: Temporal Convolution Networks}
In recent years, temporal convolution networks (e.g., FCSN \cite{ours18_eccv}) have shown to achieve impressive performance on video understanding tasks. These networks mainly perform 1D operations (e.g., 1D convolution, 1D pooling) over the temporal dimension (e.g., over frames in a video). This is analogous to the 2D operations (e.g., 2D convolution, 2D pooling) commonly used in CNN models for image-related tasks. For example, the work in \cite{ours18_eccv} uses temporal convolutional networks for video summarization, where the task is formulated as predicting a binary label for each frame in a video. Our proposed model is based on the temporal convolution network proposed in \cite{ours18_eccv}. But we extend the model to perform user-specific video highlight detection.

\subsection{Temporal-Adaptive Instance Normalization}
Let $\textbf{o}^i$ indicate the activations of $i$-th layer in the temporal convolution neural network ($f_T$) for the input video $\textbf{v}$. We use $C^i$ and $T^i$ to denote the number of channels and temporal length of activation in that layer, respectively. We define the \emph{Temporal-Adaptive Instance Normalization} (T-AIN), a conditional normalization layer for videos. T-AIN is inspired by the basic principles of Instance Normalization \cite{ulyanov17_cvpr}. The activation is firstly normalized channel-wise along \emph{temporal} dimension (obtaining $\textbf{o}_{norm}^i$), followed by a uniform modulation with affine transformation. Different from InstanceNorm \cite{ulyanov17_cvpr}, the affine parameters, scale and bias, in T-AIN are \emph{not learnable} but inferred using external data (i.e., a user's history ($\mathcal{H}$) in our case) which is encoded to a vector $m$ using another temporal convolution network ($g_T$). Thus, T-AIN is also \emph{conditional} (on $\mathcal{H}$) in nature. The activation value from T-AIN at location $c \in C^i$ and $t \in T^i$ is
\begin{equation}
\left( \frac{o^i_{c,t} - \mathrm{E}[o^i_{c}]}{ \sqrt{\mathrm{Var}[o^i_{c}] + \epsilon}} \right) \gamma_c^i + \delta_c^i, 
\label{eq:t-adain}
\end{equation}   
where $o^i_{c,t}$, $\mathrm{E}[o^i_{c}]$ and $\mathrm{Var}[o^i_{c}]$ are the activation before normalization, expectation and variance of the activation $\textbf{o}^i$ in channel $c$, respectively. T-AIN computes the $\mathrm{E}[o^i_{c}]$ and $\mathrm{Var}[o^i_{c}]$ along temporal dimension independently for each channel and every input sample (video) as:
\begin{equation}
\mathrm{E}[o^i_{c}] = \mu_{c}^i= \frac{1}{T^i}\big(\sum_{t}o^i_{c,t}\big) ,
\label{eq:t-adain-mean}
\end{equation}
\begin{equation}
\mathrm{Var}[o^i_{c}] = \mathrm{E}[(o^i_{c}-\mathrm{E}[o^i_{c}])^2]
=\frac{1}{T^i}\sum_{t}\big(o^i_{c,t}-\mu_{c}^i\big)^2 .
\label{eq:t-adain-var}
\end{equation}

\begin{figure*}
	\center
	\includegraphics[width=0.6\textwidth]{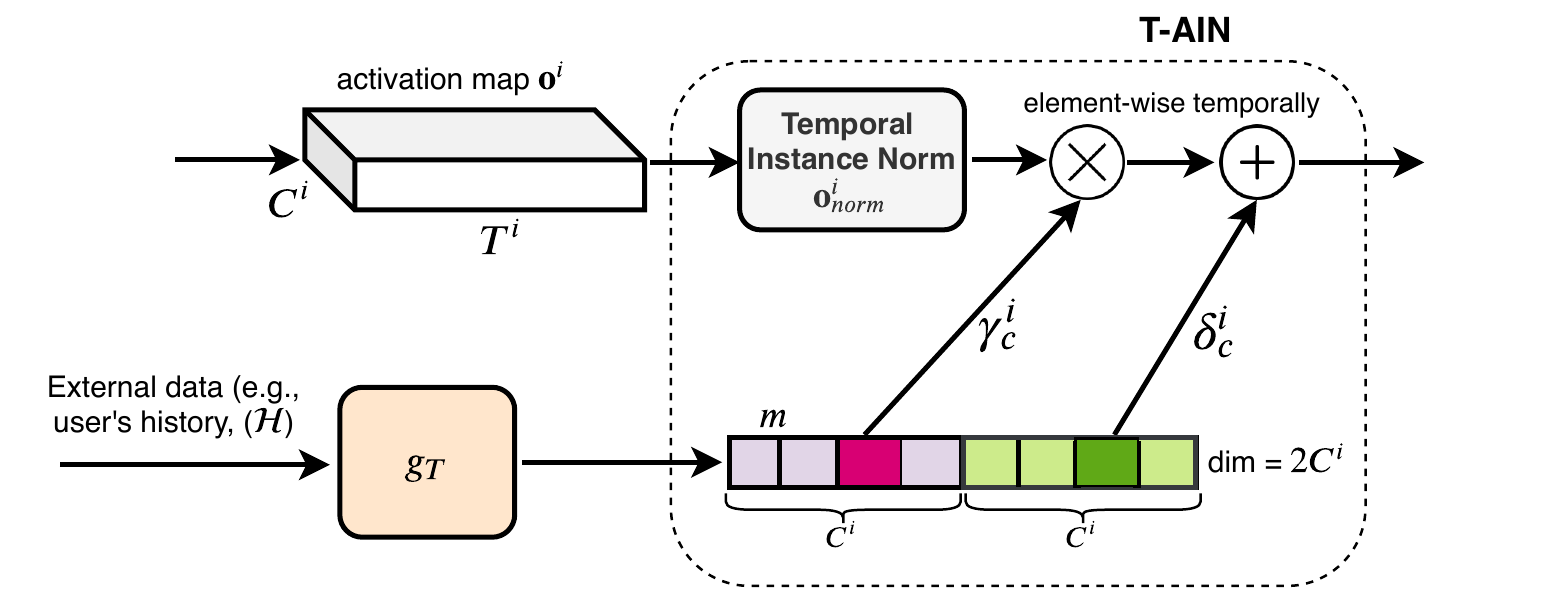}
	\caption{Overview of a temporal-adaptive instance normalization layer (T-AIN). For an input video $\textbf{v}$, let $\textbf{o}^i$ be the activation map with channel dimension $C^i$ and temporal length $T^i$ in the $i$-th layer of a temporal convolutional network $f_T$. Let $g_T$ be another temporal convolutional network that encodes external data (e.g., user's history $\mathcal{H}$) into a vector representation $m$ of dimension $2C^i$. T-AIN firstly temporally normalizes $\textbf{o}^i$ in each channel to obtain $\textbf{o}_{norm}^i$. It then uniformly scales and shifts $\textbf{o}_{norm}^i$ in channel $c$ (where $c \in C^i$) over time by $\gamma_{c}^i$ and $\delta_{c}^i$, respectively. The values of $\gamma_{c}^i$ and $\delta_{c}^i$ are obtained from $m$. As can be seen, the main characteristics of T-AIN include temporal operation, no learnable parameters, and conditional on external data.}
	\label{fig:t-adain}
\end{figure*}

In Eq. \ref{eq:t-adain}, $\gamma_c^i$ and $\delta_c^i$ are the modulation parameters in the T-AIN layer. We obtain $\gamma_c^i$ and $\delta_c^i$ from the encoded vector $m$ generated using the external data. T-AIN firstly scales each value along the temporal length in channel $c$ of temporally normalized activations $\textbf{o}_{norm}^i$ by $\gamma_c^i$ and then shifts it by $\delta_c^i$. Similar to InstanceNorm, these statistics vary across channel $C^i$. We provide details on how we compute these parameters when using user's history $\mathcal{H}$ as an external data in Sec. \ref{sec:ada_highlight}. In Fig. \ref{fig:t-adain}, we visualize the operations in a T-AIN layer.

T-AIN is related to the conditional batch normalization (CBN)~\cite{de2017modulating} and the adaptive instance normalization (AIN)~\cite{huang17_cvpr}. The main difference is that CBN and AIN work spatially, so they are appropriate for image-related tasks like style transfer. In contrast, T-AIN is designed to operate along time, which makes it suitable for video highlight detection and video understanding tasks in general.  

\subsection{Adaptive Highlight Detector}\label{sec:ada_highlight}
The adaptive highlight detector $G$ consists of two sub-networks: a highlight detection network $H$ and a history encoder network $M$. $H$ is responsible for scoring each frame in an input video to indicate whether or not it should be included in the highlight. The role of $M$ is to firstly encode the user's history information and then guide $H$ in a manner that the generated highlight is adapted to the user's history. Next, we discuss the sub-networks design and learning in detail.\\

\noindent{\bf Highlight Detection Network.}
The highlight detection network $H$ is based on FCSN \cite{ours18_eccv}. It is an encoder-decoder style architecture which is fully convolutional in nature. One advantage of this network is that it is not restricted to fixed-length input videos. It can handle  videos with arbitrary lengths. Another advantage is that it is effective in capturing the long-term temporal dynamics among the frames of a video beneficial for video understanding tasks such as highlight detection.

$H$ accepts a video $\textbf{v}$ with feature representation of dimension $1 \times T \times D$, where $T$ is number of frames in $\textbf{v}$ and $D$ is the dimension of each frame feature vector. It produces an output of dimension $1 \times T \times 2$ indicating non-highlight and highlight scores for the $T$ frames.  

The encoder ($F_{v}$) of $H$ has a stack of seven convolutional blocks. The first five blocks (i.e., $conv$-$blk1$ to $conv$-$blk5$) consist of several temporal convolution followed by a ReLU and a temporal max pooling operations. The last two blocks ($conv6$ to $conv7$) have a temporal convolution, followed by a ReLU and a dropout layer. The encoder $F_{v}$ gives two outputs: a feature map from its last layer and a skip connection from block $conv$-$blk4$. 

\begin{figure*}[h]
	\centering
	\includegraphics[width=0.6\textwidth]{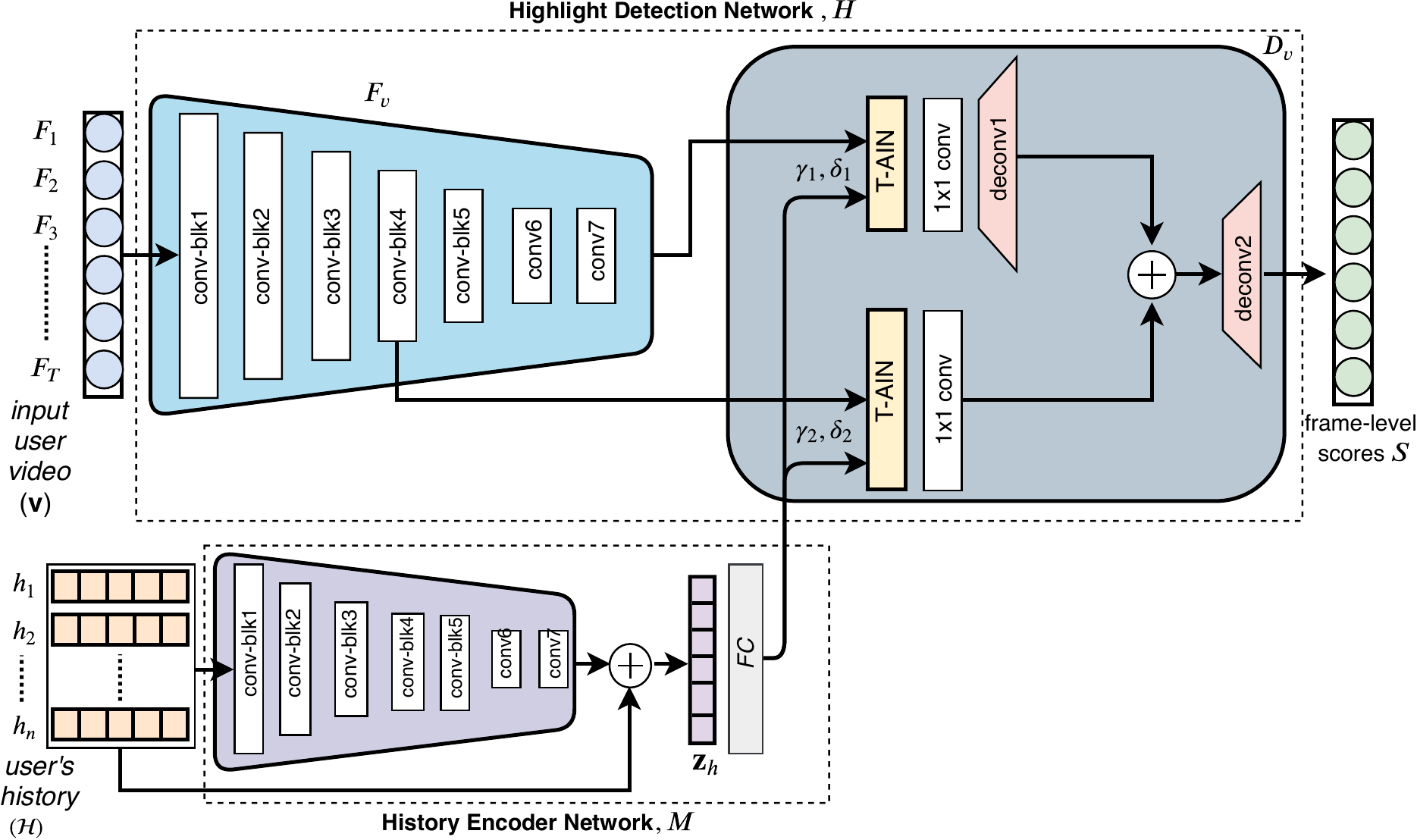}
	\caption{Overview of our proposed model, \texttt{Adaptive-H-FCSN}. The model consists of two sub-networks, a \textit{highlight detection network} $H$ and a \textit{history encoder network} $M$. $H$ is an encoder-decoder architecture that takes a frame-level vector feature representation of a user input video with $T$ frames. It then generates scores (highlight vs. non-highlight) for each frame in the video while taking information from $M$. $M$ takes vector feature representation of each element (i.e., highlights the user has previously created) in the user's history as an input and encodes it to a vector $\textbf{z}_h$. This vector $\textbf{z}_h$ is then simply fed to a fully connected layer $FC$ to produce the affine parameters $\gamma_j$ and $\delta_j$ in the $j$-th T-AIN layer of decoder $D_v$ where $j = 1, 2$. This way the highlight detection for the input video is adapted to the user.}
	\label{fig:combined-arch}
\end{figure*}

The output of encoder $F_{v}$ is fed to the decoder network ($D_{v}$). We introduce T-AIN layers at sites where these two outputs enter $D_{v}$. We obtain a feature map by applying a $1\times1$ convolution and a temporal fractionally-strided convolution operation ($deconv1$) to the output of first T-AIN, which is added with the feature map from a $1\times1$ convolution to the output of second T-AIN layer. Finally, we apply another fractionally-strided temporal convolution ($deconv2$) to obtain a final prediction of shape $1 \times T \times 2$ denoting two scores (non-highlight or highlight) for each frame in the video. Fig. \ref{fig:combined-arch} (top) visualizes the architecture of $H$.\\

\noindent{\bf History Encoder Network.}
The history encoder network $M$ is an integral piece of our framework. It acts as a guiding network that guides the highlight network $H$ by adaptively computing the affine transformation parameters of its T-AIN layers. Using these affine parameters, $M$ modulates the activations in $H$ to produce adaptive highlight predictions. 

The configuration of this network is same as the encoder $F_{v}$ in $H$ with few changes towards the end. It performs an average temporal pooling to the output of convolution blocks, which is combined with a skip connection from the input that is then fed to a fully-connected layer. The skip connection involves an average pooling and a fully-connected operation to match dimensions.

The network accepts user's history collection $\mathcal{H}$ of shape $1 \times n \times D$ as an input, where $n$ is the number of elements/highlights in the user's history and $D$ is the dimension of the vector representation of each element. In our implementation, we obtain a $D$-dimensional vector from a highlight using C3D~\cite{tran15_iccv}. Note that $n$ varies for different users. After the combining stage, the network generates a latent code $\textbf{z}_h$ which is a fixed-length user-specific vector. 

Next, we forward $\textbf{z}_h$ to a fully connected layer ($FC$) to decode the latent code $\textbf{z}_h$ into a set of vectors ($\gamma_j,\delta_j$) where $j=1, 2$. The parameters $\gamma_j$ and $\delta_j$ denote the scaling and bias parameters of $j$-th T-AIN layer in the decoder $D_{v}$ of $H$, respectively. These affine parameters are uniformly applied at every temporal location in the feature map. This allows to incorporate the user's history information in $H$ and adjust it in a way that the predicted highlight is adapted to the user's history. This way we obtain a user-specific highlight prediction for the input user video. Fig. \ref{fig:combined-arch} (bottom) presents the architecture of $M$.

With $F_v$, $D_v$ and $M$, we can rewrite Eq. \ref{eq:fn_G} as:
\begin{align}
S &= D_v(F_v(\textbf{v}), M(\{h_1,...,h_n\})).
\label{eq:fn_G_modified}
\end{align}
By applying this design, we learn a generic video representation using $F_v$ and extract a user-specific latent representation with $M$. Finally, by injecting user-specific latent information to $D_v$ through T-AIN layers, we allow the model to adapt the highlight detection to the user's history.

Note that we use temporal convolutions over the highlights in the user history. In addition, we also investigate a non-local model \cite{wang2018_cvpr,zhang2019_icml} with self-attention instead of temporal convolutions for $M$. Here, the output of self-attention is firstly average pooled to produce a single vector and then fed to a fully-connected layer. We find that the non-local model performs slightly inferior to the temporal convolutions model. This is probably because the highlights in the history are ordered based on their time of creation in the dataset, so the temporal convolutions allow the history encoder $M$ to capture some implicit temporal correlations among them.

\subsection{Learning and Optimization}
We train our adaptive highlight detector $G$ using a cross-entropy loss. For an input video $\textbf{v}$ with $T$ frames and corresponding binary indicator ground truth label vector (indicating whether a frame is non-highlight or highlight), we define a highlight detection loss $\mathcal{L}_{highlight}$ as:
\begin{equation}\label{eq:loss}
\mathcal{L}_{highlight} = -\frac{1}{T}\sum_{t=1}^{T} \log\Bigg(\frac{\exp(\lambda_{t,l_{c}})}{\sum_{c=1}^{2}\exp(\lambda_{t,c})}\Bigg) ,
\end{equation}
where $\lambda_{t,c}$ is the predicted score of $t$-th frame to be the $c$-th class (non-highlight or highlight) and $\lambda_{t,l_c}$ is the score predicted for the ground truth class. 

The goal of our learning is to find optimal parameters $\Theta_{F_v}^*$, $\Theta_{D_v}^*$ and $\Theta_{M}^*$ in the encoder $F_v$, decoder $D_v$ of the highlight detection network $H$, and the history encoder network $M$, respectively. The learning objective can be expressed as:
\begin{equation}\label{eq:goal}
\Theta_{F_v}^*,\Theta_{D_v}^*,\Theta_{M}^* =  \underset{\Theta_{F_v},\Theta_{D_v},\Theta_{M}}{\arg\min} \; \mathcal{L}_{highlight}(F_v, D_v, M) .
\end{equation}

For brevity, we use \texttt{Adaptive-H-FCSN} to denote our adaptive highlight detection model learned using Eq. \ref{eq:goal}.

%% file: experiments.tex
\section{Experiments}\label{sec:exp}

\subsection{Dataset}\label{sec:exp_dataset}
We conduct experiments on the largest publicly available highlight detection dataset, PHD-GIFs \cite{gygli18_acmmm}. It is also the only large-scale dataset that has user history information for highlight detection. The released dataset consists of $119,938$ videos, $13,822$ users and $222,015$ annotations. The dataset has $11,972$ users in training, $1,000$ users in validation, and $850$ users in testing. There is no overlap among users in these three subsets.

Apart from being large-scale, this dataset is also interesting because it contains user-specific highlight examples indicating what exactly a user is interested in when creating highlights. The ground truth segment-level annotation comes from GIFs that a user creates (by extracting key moments) from YouTube videos. In this dataset, a user has GIFs from multiple videos where the last video of the user is considered for highlight prediction and the remaining ones are treated as examples in the user's history.

The dataset only provides YouTube video ID for the videos and not the original videos. So we need to download the original videos from YouTube to carry out the experiments. We were able to download $104,828$ videos and miss the remaining videos of the dataset since they are no longer available on YouTube. In the end, we are able to successfully process $7,036$ users for training, $782$ users for validation and $727$ users for testing. Note that code of previous methods on this dataset are not available (except pre-trained Video2GIF \cite{gygli16_cvpr}), so we implement several strong baselines (see Sec. \ref{sec:exp_baselines}) in the paper.

\subsection{Setup and Implementation Details}\label{sec:exp_setup}
\noindent{\bf Evaluation metrics}: We use the mean Average Precision (mAP) as our evaluation metric. It measures the mean of the average precision of highlight detection calculated for every video in the testing dataset. Different from object detection where all the detections are accumulated from images to compute the average precision, highlight detection treats videos separately because it is not necessary a highlighted moment in a particular video is more interesting than non-highlighted moments in a different video~\cite{sun2014ranking}. This metric is commonly used to measure the quality of predictions in highlight detection~\cite{gygli16_cvpr,gygli18_acmmm,sun2014ranking,xiong19_cvpr}.

\noindent{\bf Feature representation}: Following prior work \cite{gygli18_acmmm}, we extract C3D \cite{tran15_iccv} (conv5) layer features and use it as feature representation in the model for the input videos and user's history. For an input video, we extract C3D-features at frame-level. For a highlight video in the user's history, we prepare a single vector representation by averaging its frame-level C3D features. 

\noindent{\bf Training details}: We train our models from scratch. All the models are trained with a constant learning rate of 0.0001. We use Adam  \cite{kingma2014adam} optimization algorithm for training the models. Note that we apply this training strategy in all our experiments including the other analysis (Sec. \ref{sec:ablation}).

Since the dataset has users that create multiple GIFs for a video, we follow \cite{gygli18_acmmm} to prepare a single ground truth for the video by taking their union.

\noindent{\bf Testing details}: Given a new test user video and the user's history, we use our trained model to predict highlight score for each frame which is then sent to the evaluation metrics to measure the quality of predicted highlight. We follow the evaluation protocol of previous work \cite{gygli16_cvpr,gygli18_acmmm} for fair comparison. Note that our model can handle variable length input videos and variable number of history elements. We consider full user's history while predicting highlights.

\subsection{Baselines}\label{sec:exp_baselines}
We compare the proposed \texttt{Adaptive-H-FCSN} with the following strong baselines:

\textbf{FCSN}~\cite{ours18_eccv}: This network is the state of the art in video summarization which we adapt as our highlight detection network. FCSN has no instance normalization layers. We train and evaluate FCSN on the PHD-GIFs dataset.

\textbf{Video2GIF}~\cite{gygli16_cvpr}: This baseline is a state-of-the-art highlight detection model. We evaluate the publicly available pre-trained model.    

\textbf{FCSN-aggregate}: In this baseline, we train FCSN~\cite{ours18_eccv} by directly combining the user history with the input video. More specifically, we firstly obtain a vector representation for the user history by averaging the features of elements in the history. Next, we add this aggregated history with the feature representation of each frame in the input video.

\textbf{H-FCSN}: This baseline is a variant of highlight detection network $H$ we presented in Sec. \ref{sec:ada_highlight}, where we replace the T-AIN layers in the decoder of $H$ with the unconditional temporal instance normalization layers. We do not have the history encoder network $M$. This results in \texttt{Adaptive-H-FCSN} transformed to a generic highlight detection model with no adaptation to users.

\textbf{H-FCSN-aggregate}: Similar to FCSN-aggregate, we directly combine the user's history features with an input video features and learn H-FCSN. Different from H-FCSN, this is not a generic highlight detection model but a user-adaptive highlight detection model as we allow the model to leverage the user's history information in the training and inference. 

\subsection{Results and Comparison}\label{sec:exp_results}
In Table \ref{table:main_results}, we provide the experiment results (in terms of mAP \%) of our final model \texttt{Adaptive-H-FCSN} along with the baselines and other alternative methods. 

\begin{table*}
	\centering
	\caption{Performance (mAP\%) comparison between \texttt{Adaptive-H-FCSN} and other approaches. We compare with both non-adaptive and adaptive highlight detection methods. Our method \texttt{Adaptive-H-FCSN} outperforms the other alternative methods. We also compare with \texttt{Adaptive-H-FCSN-attn} that uses self-attention in the history encoder (see Sec. \ref{sec:ada_highlight}). Note that all the listed methods use C3D feature representation}
	\begin{tabular}{c|c|c}
	\hline
	Method & mAP (\%) & User-adaptive \\
	\hline
	Random & 12.27 &  \xmark\\
	FCSN \cite{ours18_eccv} & 15.22 & \xmark\\
	Video2GIF \cite{gygli16_cvpr} & 14.75 &  \xmark\\
	H-FCSN & 15.04  & \xmark  \\
	FCSN-aggregate & 15.62 & \cmark\\
	H-FCSN-aggregate & 15.73  & \cmark  \\
	\hline
	\texttt{Adaptive-H-FCSN-attn} & 16.37 & \cmark  \\
	\texttt{Adaptive-H-FCSN} & \textbf{16.73} & \cmark  \\
	\hline
	\end{tabular}
	\label{table:main_results}
\end{table*}

\texttt{Adaptive-H-FCSN} outperforms all the baselines. Results show that directly combining (i.e., FCSN-aggregate and H-FCSN-aggregate) history information with input video only slightly improves the highlight detection results in comparison to FCSN and H-FCSN that do not leverage users history information. However, we notice a significant performance gain for \texttt{Adaptive-H-FCSN} model. This result validates that directly combining user history information with the input video is a sub-optimal solution for user-adaptive highlight detection. Additionally, this result reveals that proposed T-AIN layer plays a critical role in producing more accurate and user-specific highlight detection. It is also noteworthy that we obtain a lower performance (nearly 1\%) for Video2GIF~\cite{gygli16_cvpr} than reported in PHD-GIFs~\cite{gygli18_acmmm} which implies that our test set is more challenging.

\begin{figure*}[h]
	\center
	\includegraphics[width=0.7\textwidth]{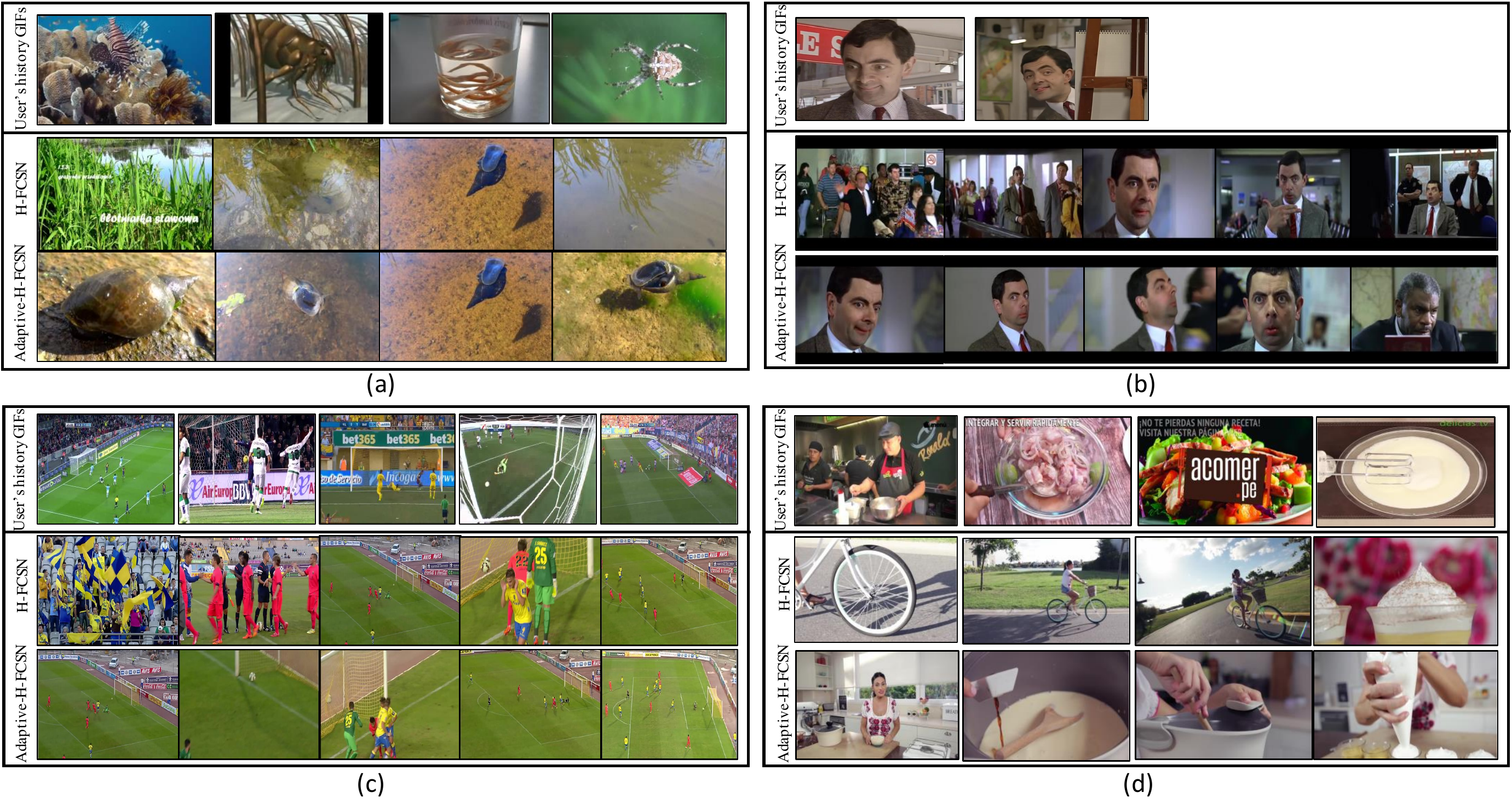}
	\caption{Qualitative examples for different methods. We show examples of the generic highlight detection model (H-FCSN) and our user-adaptive model (\texttt{Adaptive-H-FCSN}) on four videos. For each video, we show the user's history (multiple GIFs) and few sampled frames from the highlight predictions of the two models. Based on the user's history, we find that in (a) the user has interest in animals; (b) the user is interested in faces that dominate a scene; (c) the user is inclined to highlight goal scoring scenes; and (d) the user focuses on cooking. These visualizations indicate that adaptation to the user's interest is important for a meaningful and accurate highlights. Compared with H-FCSN, the prediction of \texttt{Adaptive-H-FCSN} is more consistent with the user history. }
	\label{fig:qual}
\end{figure*}

Fig. \ref{fig:qual} shows some qualitative examples for the generic baseline model (H-FCSN) and our proposed adaptive highlight detection model (\texttt{Adaptive-H-FCSN}). We can see that our model successfully captures the information in user's history and produces highlights that are adapted to the user's preference.

\subsection{Analysis}\label{sec:ablation}

\subsubsection{Effect of affine parameters.} We analyze the importance of affine parameters $\gamma_c^i$ and $\delta_c^i$ (Eq. \ref{eq:t-adain}) for adaptive highlight detection. In Table \ref{table:affine_parames}, we report the highlight detection performance for different possible choices of these parameters. We find that when these parameters are adaptively computed ($\gamma_c^i$=$\gamma_c^h$, $\delta_c^i$=$\delta_c^h$) from another network capturing the user history information (i.e., \texttt{Adaptive-H-FCSN}) significantly boosts the highlight detection performance as opposed to cases when it is directly learned ($\gamma_c^i$=$\gamma_c^*$, $\delta_c^i$=$\delta_c^*$) and set to a fixed value ($\gamma_c^i$=1, $\delta_c^i$=0) in the main highlight detection network. Thus, the proposed T-AIN layer is key to obtain user-adaptive highlights.
\begin{table}
	\centering
	\caption{Impact of affine parameters on highlight detection. Here we show the  performance (mAP\%) for different choices of affine parameters $\gamma_c^i$ and $\delta_c^i$ in Eq. \ref{eq:t-adain}}
		\begin{tabular}{c|c|c|c}
			\hline
			 Method & $\gamma_c^i$=1, $\delta_c^i$=0  & $\gamma_c^i$=$\gamma_c^*$, $\delta_c^i$=$\delta_c^*$ & $\gamma_c^i$=$\gamma_c^h$, $\delta_c^i$=$\delta_c^h$\\
			 \hline
			H-FCSN  & 14.64  & 15.04 & - \\
			H-FCSN-aggregate  & 14.87  & 15.73 & - \\
			\texttt{Adaptive-H-FCSN} & -  & - & \textbf{16.73} \\
			\hline
		\end{tabular}
	\label{table:affine_parames}
\end{table}

\subsubsection{Effect of user's history size.} We perform additional study to analyze how sensitive our model is to the length of a user's history (i.e., numbers of highlights previously created). We restrict the number of history elements for users in the training. That is, we consider only $h$ highlight videos from the user's history in training. During testing, we consider the user's full history.

Table \ref{table:hist_size} shows the results of various methods as a function of number of elements ($h$ = $0, 1, 5, n$) in user's history. We observe that \texttt{Adaptive-H-FCSN} outperforms generic highlight model (H-FCSN) even when there is a single highlight in the user's history. We notice the performance of \texttt{Adaptive-H-FCSN} gradually improves when we increase the number of history elements, whereas H-FCSN-aggregate doesn't show similar performance trend. It achieves the best results when we utilize a user's full history (i.e., $h$=$n$).

\begin{table}
	\centering
	\caption{Impact of an user's history size (i.e., number of history elements/highlights) on different methods. Here we vary the history size $h$ as 0 (no history), 1, 5, and $n$ (full history). The performance of our model improves with the increase in history size}
		\begin{tabular}{c|c|c|c|c}
			\hline
			History size ($\mathcal{H}$) & $h=0$ & $h=1$ & $h=5$ & $h=n$\\
			\hline
			H-FCSN & 15.04 & - & - & - \\
			H-FCSN-aggregate & - & 15.62  & 15.04  & 15.73 \\
			\texttt{Adaptive-H-FCSN} & - & 15.57 & 15.69 & \textbf{16.73}\\
			\hline
		\end{tabular}
	\label{table:hist_size}
\end{table}

%% file: vsumm.tex
\subsection{Application to Video Summarization}\label{sec:vsumm}
Video summarization is closely related to highlight detection. Highlight detection aims at extracting interesting moments and events of a video, while video summarization aims to generate a concise synopsis of a video. Popular datasets in summarization are very small \cite{ours19_cvpr}, making learning and optimization challenging. We argue that pretraining using a large-scale video data from a related task, such as PHD-GIFs \cite{gygli18_acmmm} in highlight detection, could tremendously help video summarization models. In video summarization, this idea remains unexplored. In order to validate our notion and compare with recent state-of-the-art in \cite{wei2018sequence}, we select the SumMe dataset \cite{gygli14_eccv} which has only $25$ videos.  

We evaluate our trained H-FCSN (i.e., the generic highlight detection model we proposed in Sec. \ref{sec:exp_baselines}) directly on SumMe. In Table \ref{table:summe}, we compare the performance of our H-FCSN (trained on the PHD-GIFs \cite{gygli18_acmmm} dataset) on SumMe with state-of-the-art supervised video summarization methods. Following prior work \cite{wei2018sequence}, we randomly select $20\%$ of data in SumMe for testing. We repeat this experiment five times (as in \cite{wei2018sequence}) and report the average performance. Surprisingly, even though we \textit{do not} train on SumMe, our model achieves state-of-the-art summarization performance, outperforming contemporary supervised models. Therefore, we believe that the future research in video summarization should consider pretraining their model on a large-scale video data from a related task such as highlight detection. We envision that this way we can simultaneously make progress in both highlight detection and video summarization.
\begin{table}
	\centering
	\caption{Performance comparison in term of F-score (\%) on SumMe. Note that unlike other methods, we do not train on SumMe rather directly test our trained (using PHD-GIFs) model for summarization. Results of other methods are taken from \cite{wei2018sequence}}
		\begin{tabular}{|c|c|c|}
			\hline
			Method & F-score (\%)\\
			\hline
			Interesting \cite{gygli14_eccv} & 39.4\\
			Submodularity \cite{gygli15_cvpr} & 39.7\\
			DPP-LSTM \cite{zhang16_eccv} & 38.6\\
			GAN$_{sup}$ \cite{mahasseni17_cvpr} & 41.7\\
			DR-DSN$_{sup}$ \cite{zhou2018_aaai} & 42.1\\
			S$^2$N \cite{wei2018sequence} & 43.3\\
			\hline
			Ours (H-FCSN) & \textbf{44.4}\\
			\hline
		\end{tabular}
	\label{table:summe}
\end{table}

%% file: conclusion.tex
\section{Conclusion}\label{sec:conclude}
We have proposed a simple yet novel framework \texttt{Adaptive-H-FCSN} for adaptive highlight detection using the user history which has received less attention in the literature. Different from commonly applied ranking model, we introduced a convolutional model for highlight detection that is computationally efficient as it can process an entire video of any length at once and also does not require expensive shot detection computation. We proposed temporal-adaptive normalization (T-AIN) that has affine parameters which is adaptively computed using the user history information. The proposed T-AIN leads to high-performing and user-specific highlight detection. Our empirical results on a large-scale dataset indicate that the proposed framework outperforms alternative approaches. Lastly, we further demonstrate an application of our learned model in video summarization where the learning is currently limited to small datasets.\\

\noindent \textbf{Acknowledgements.} The work was supported by NSERC. We thank NVIDIA for donating some of the GPUs used in this work.